\def\year{2021}\relax
\documentclass[letterpaper]{article} 
\usepackage{aaai21}  
\usepackage{times}  
\usepackage{helvet} 
\usepackage{courier}  
\usepackage[hyphens]{url}  
\usepackage{graphicx} 
\usepackage{natbib}  
\usepackage{caption} 
\usepackage{booktabs}
\usepackage{amsmath}
\usepackage{algorithmic}
\usepackage{textcomp}
\usepackage{xcolor}
\usepackage{fancyhdr}
\usepackage{float}

\title{STAR: \underline{S}parse \underline{T}ransformer-based \underline{A}ction \underline{R}ecognition}
\author{
    
    Feng Shi\textsuperscript{\rm 1}, 
    Chonghan Lee\textsuperscript{\rm 2}, 
    Liang Qiu\textsuperscript{\rm 1}, 
    Yizhou Zhao\textsuperscript{\rm 1}, 
    Tianyi Shen\textsuperscript{\rm 2}, 
    Shivran Muralidhar\textsuperscript{\rm 2}, 
    Tian Han\textsuperscript{\rm 3},
    Song-Chun Zhu\textsuperscript{\rm 1}, 
    Vijaykrishnan Narayanan\textsuperscript{\rm 2}
    \\
}
\affiliations{
    \textsuperscript{\rm 1}University of California Los Angeles\\

    \textsuperscript{\rm 2}The Pennsylvania State University\\
    \textsuperscript{\rm 3}Stevens Institute of Technology\\

}

\begin{document}

\maketitle

\begin{abstract}
The cognitive system for human action and behavior has evolved into a deep learning regime, and especially the advent of Graph Convolution Networks has transformed the field in recent years. However, previous works have mainly focused on over-parameterized and complex models based on dense graph convolution networks, resulting in low efficiency in training and inference. Meanwhile, the Transformer architecture-based model has not yet been well explored for cognitive application in human action and behavior estimation. This work proposes a novel skeleton-based human action recognition model with sparse-attention on the spatial dimension and segmented linear attention on the temporal dimension of data. Our model can also process variable length of video clips grouped as a single batch. Experiments show that our model can achieve comparable performance while utilizing much less trainable parameters and achieve high speed in training and inference. Experiments show that our model achieves 4$\sim$18$\times$ speedup and $\frac{1}{7}$$\sim$$\frac{1}{15}$ model size compared with the baseline models at competitive accuracy.
\end{abstract}

\section{Introduction}
Human action recognition plays a crucial role in many real-world applications, such as holistic scene understanding, video surveillance, and human-computer interaction \cite{vsurveilance, hcinter/978-3-540-78566-8_10}. In particular, skeleton-based human action recognition has attracted much attention in recent years and has shown its effectiveness. The skeleton representation contains a time series of 2D or 3D coordinates of human key-joints, providing dynamic body movement information that is robust to variations of light conditions and background noises in contrast to raw RGB representation.

Earlier skeleton-based human action recognition methods focus on designing hand-crafted features extracted from the joint coordinates \cite{lie2014, actionlet2012} and aggregating learned features using RNNs and CNNs \cite{rnndu, xiememory, zhang2017view, ts3dliu, tdskrepke}. However, these methods rarely explore the relations between body joints and result in unsatisfactory performance. Recent methods focus on exploring the natural connection of human body joints and successfully adopted the Graph Convolutional Networks (GCNs), especially for non-Euclidean domain data, similar to Convolutional Neural Networks (CNNs) but executing convolutional operations to aggregate the connected and related joints' features. Yan et al. \cite{yan2018spatial} proposed a ST-GCN model to extract discriminative features from spatial and temporal graphs of body joints. Following the success of ST-GCN, many works proposed optimizations to ST-GCN to improve the performance and network capacity \cite{directed2019, Li_2019_CVPR, liu2020disentangling}.

However, the existing GCN-based models are often impractical in real-time applications due to their vast computational complexity and memory usage. The baseline GCN model, e.g., ST-GCN, consists of more than 3.09 million parameters and costs at least 16.2 GFLOPs to run inference on a single action video sample \cite{yan2018spatial}. DGNN, which is composed of incremental GCN modules, even contains 26 million model parameters. \cite{directed2019} Such high model complexity leads to difficulty in model training and inference, makes the model not suitable for deployment on edge devices. Furthermore, these GCN-based models process fixed-size action videos by padding repetitive frames and zeros to match the maximum number of frames and persons depicted in the videos. These additional paddings increase the latency and memory required hindering their adoption in real-time and embedded applications.

This paper proposes a \textit{sparse transformer-based action recognition} (STAR) model as a novel baseline for skeleton action modeling to address the above shortcomings. Transformers have been a popular choice in natural language processing. Recently, they have been employed in computer vision to attain competitive results compared to convolutional networks, while requiring fewer computational resources to train \cite{dosovitskiy2020image, huang2019ccnet}. Inspired by these Transformer architectures, our model consists of spatial and temporal encoders, which apply sparse attention and segmented linear attention on skeleton sequences along the spatial and temporal dimension, respectively.
Our sparse attention module along the spatial dimension performs sparse matrix multiplications to extract correlations of connected joints, whereas previous approaches utilize dense matrix multiplications where most of the entries are zeros, causing extra computation. The segmented linear attention mechanism along temporal dimension further reduces the computation and memory usage by processing variable length of sequences. We also apply segmented positional encoding to the data embedding to provide the concept of time-series ordering along the temporal dimension of variable-length skeleton data. Additionally, segmented context attention performs weighted summarization across the entire video frames, making our model robust compared to GCN-based models with their fixed-length receptive field on the temporal dimension.

Compared to the baseline GCN model (ST-GCN), our model (STAR) achieves higher performance with much smaller model size on the two datasets, NTU RGB+D 60 and 120. The major contributions of this work are listed below:
\begin{itemize}
    \item We focus on designing an efficient model purely based on self-attention mechanism. We propose \textit{sparse transformer-based action recognition} (STAR) model that process variable length of skeleton action sequence without additional preprocessing and zero paddings. The flexibility of our model is beneficial for real-time applications or edge platforms with limited computational resources.
    \item  We propose a sparse self-attention module that efficiently performs sparse matrix multiplications to capture spatial correlations between human skeleton joints.
    \item We propose a segmented linear self-attention module that effectively captures temporal correlations of dynamic joint movements across time dimension.
    \item Experiments show that our model is 5$\sim$7$\times$ smaller than the baseline models while providing 4$\sim$18$\times$ execution speedup.
\end{itemize}

\section{Related works}
\subsection{Skeleton-Based Action Recognition}
Recently, skeleton-based action recognition has attracted much attention since its compact skeleton data representation makes the models more efficient and free from the variations in lighting conditions and other environmental noises. Earlier methods to skeleton-based action modeling have mainly worked on designing hand-crafted features and relations between joints \cite{}. 
Recently, by looking into the inherent connectivity of the human body, Graph Convolutional Networks (GCNs), especially, ST-GCNs have gained massive success in getting satisfactory results in this task. The model consists of spatial and temporal convolution modules similar to conventional convolutional filters used for images \cite{yan2018spatial}. The graph adjacency matrix encodes the skeleton joints' connections and extracts high-level spatial representations from the skeleton action sequence. On the temporal dimension, 1D convolutional filters facilitate extracting dynamic information.

Many following works have proposed improvements to ST-GCN to improve the performance. Li et al. \cite{Li_2019_CVPR} proposed AS-GCN, which leveraged the potential of adjacency matrices to scale the human skeleton's connectivity. Furthermore, they generated semantic links to capture better structural and action semantics with additional information aggregation. Lei et al. \cite{shi2019skeleton} proposed Directed Graph Neural Networks (DGNNs), which incorporate joint and bone information to represent the skeleton data as a directed acyclic graph. Liu et al. \cite{liu2020disentangling} proposed a unified spatial-temporal graph convolution module (G3D) to aggregate information across space and time for effective feature learning.

Some studies have been focusing on the computational complexity of GCN-based methods. Cheng et al. \cite{cheng2020shiftgcn} proposed Shift-GCN, which leverages shift graph operations and point-wise convolutions to reduce the computational complexity. Song et al. \cite{song2020stronger} proposed multi-branch ResGCN that fuses different spatio-temporal features from multiple branches and used residual bottleneck modules to obtain competitive performance with less number of parameters. Compared to these methods, our spatial and temporal self-attention modules have several essential distinctions: our model can process variable length of skeleton sequence without preprocessing with zero-paddings. Our model can retrieve global context on the temporal dimension by applying self-attention to the input sequence's entire frames.

\subsection{Transformers and Self-Attention Mechanism}
Vaswani et al. \cite{attn2017all} first introduced Transformers for machine translation and have been the state-of-the-art method in various NLP tasks. For example, GPT and BERT \cite{radford2018improving, devlin-etal-2019-bert} are currently the Transformer-based language models that have achieved the best performance. The core component of Transformer architectures is a self-attention mechanism that learns the relationships between each element of a sequence. In contrast to recurrent networks that process sequence in a recursive fashion and are limited to attention on short-term context, transformer architectures enable modeling long dependencies in sequence. Furthermore, the multi-head self-attention operations can be easily parallelized. Recently, Transformer-based models have attracted much attention in the computer vision community. Convolution operation has been the core of the conventional deep learning models for computer vision tasks. However, there are downfalls to the operation. The convolution operates on a fixed-sized window, which only captures short-range dependencies. The same applies to GCNs where the Graph Convolution operation is incapable of capturing long-range relations between joints in both spatial and temporal dimensions. 

Vision Transformer (ViT) \cite{dosovitskiy2020image} is the first work to completely replace standard convolutions in deep neural networks on large-scale image recognition tasks. Huang et al. \cite{huang2019ccnet} explored the sparse attention to study the trade-off between computational efficiency and performance of a Transformer model on the image classification task. A recent study \cite{plizzari2020spatial} proposed a hybrid model consists of the Transformer encoder and GCN modules on the skeleton-based human action recognition task. Nevertheless, no prior study has completely replaced GCNs with the Transformer architecture to the best of our knowledge.

\section{Methodology}
 
In this section, we present the algorithms used in our model and the relevant architecture of our model. 

Section \ref{sec:sparse_attn} depicts the \textbf{sparse multi-head self-attention} (MHSA) mechanism used in spatial Transformer encoder module; Section \ref{sec:temp_attn} introduces the novel data format and the relevant \textbf{linear multi-head self-attention} (MHSA) mechanism for temporal Transformer encoder; Section \ref{sec:arch} shows the overall framework of our model and related auxiliary modules.

\begin{figure}[ht]
    \centering
    \includegraphics[width=0.3\textwidth]{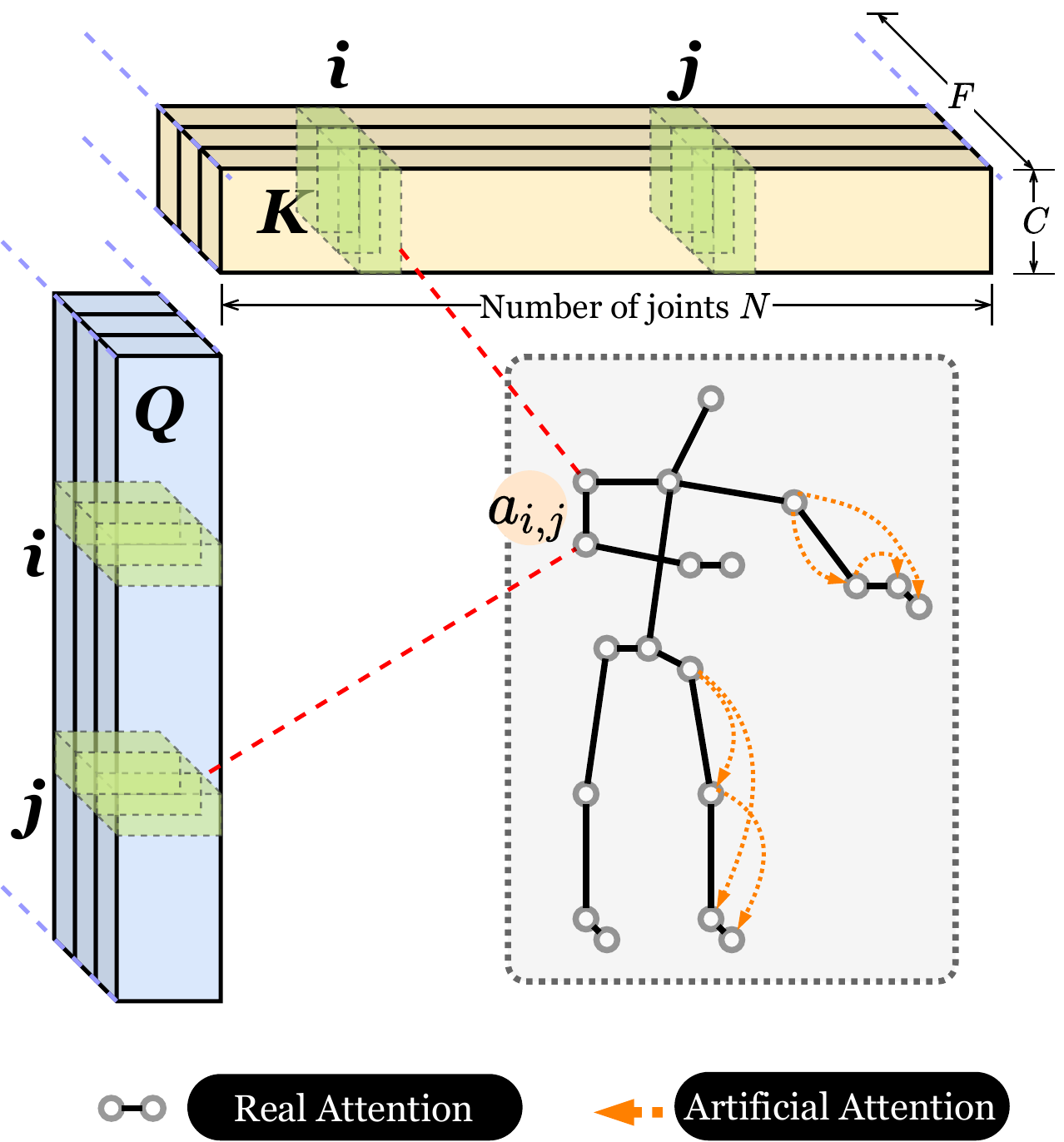}
    \caption{Illustration of our Sparse attention module: Given the queries $Q$ and the keys $K$ of the skeleton embedding, feature vectors of joint $i$ and $j$ are correlated with attention weight $\alpha_{i,j}$ The solid black line on the skeleton represents the physical connection of human skeleton. The dashed line connecting two joints represents the artificial attention of joints.}
    \label{fig:sparse_attn}
\end{figure}

\subsection{Spatial domain: Sparse MHSA} \label{sec:sparse_attn}

The crucial component of our \textit{spatial transformer encoder} is the \textit{sparse multi-head self-attention} module. GCN-based models and previous Transformer models, such as ST-GCN and ST-TR, utilize dense skeleton representation to aggregate the features of neighboring nodes. This dense adjacency matrix representation contains 625 entries for the NTU dataset, while the actual number of joint connections representing the skeletons is only 24. It means that 96\% of the matrix multiplications are unnecessary calculations for zero entries. So we propose a sparse attention mechanism, which only performs matrix multiplications on the sparse node connections. This allows each joint to only aggregate the information from its neighboring joints based on the attention coefficients, which are dynamically assigned to the corresponding connections.

The joint connections are based on the topology of skeleton, which is a tree structure. The attentions inherited from this topology are seen as \textit{physical attention} (or \textit{real attention}), as illustrated in Figure \ref{fig:sparse_attn}. To augment the attending field, we also artificially add more links between joints according to the logical relations of body parts, and we call these artificially created attentions as \textit{artificial attention}, as the dashed yellow arrows shown in Figure \ref{fig:sparse_attn}. For simplicity, suppose that the skeleton adjacency matrix is $A$, then the artificial links for additional spatial attention are obtained through $A^2$ and $A^3$. Hence, in our model, the spatial attention maps are evaluated based on the topology representation of $A + A^2 + A^3$.

The sparse attention is calculated according to the connectivity between joints. As described in below equations: after the embedding in Equation \ref{eq:embedding}, the joint-to-joint attention between a pair of connected joints is computed first by an exponential score of the dot product of the feature vectors of these two joints (Equation \ref{eq:attn}), then the score is normalized by the sum of exponential scores of all neighboring joints as described in Equation \ref{eq:sparse_attn}.
\begin{align}
    Q &= X W_q, K = X W_k, V = X W_v \label{eq:embedding}\\
    \alpha_{i, j} &= \frac{\left<q_i, k_j\right>}{\sum_{n \in N(i)} \left< q_i, k_n \right>}
   \label{eq:attn} \\
    v_{i}' &= \sum_{j \in N(i)} \alpha_{i, j} v_j, \quad\text{ or }  V' = \mathcal{A} V
    \label{eq:sparse_attn}
\end{align}

where $Q$, $K$, and $V$ are queries, keys, and values in Transformer's terminology, respectively; and $q_i = Q(i)$, $k_j = K(j)$, $v_j = V(j)$, and $\left< q, k \right> = exp\left( \frac{q^T k}{\sqrt{d}} \right)$. Finally, we obtain attention maps $\mathcal{A}$ as multi-dimension (multi-head) sparse matrices sharing the identical topology described by a single adjacency matrix (including links for the artificial attention), where attention coefficients are $\mathcal{A}(i, j) = \alpha_{i, j}$. The sparse operation can be fulfilled with tensor gathering and scattering operations for parallelism.

\subsection{Temporal domain: Segmented Linear MHSA} \label{sec:temp_attn}
\begin{figure}[ht]
    \centering
    \includegraphics[width=0.4\textwidth]{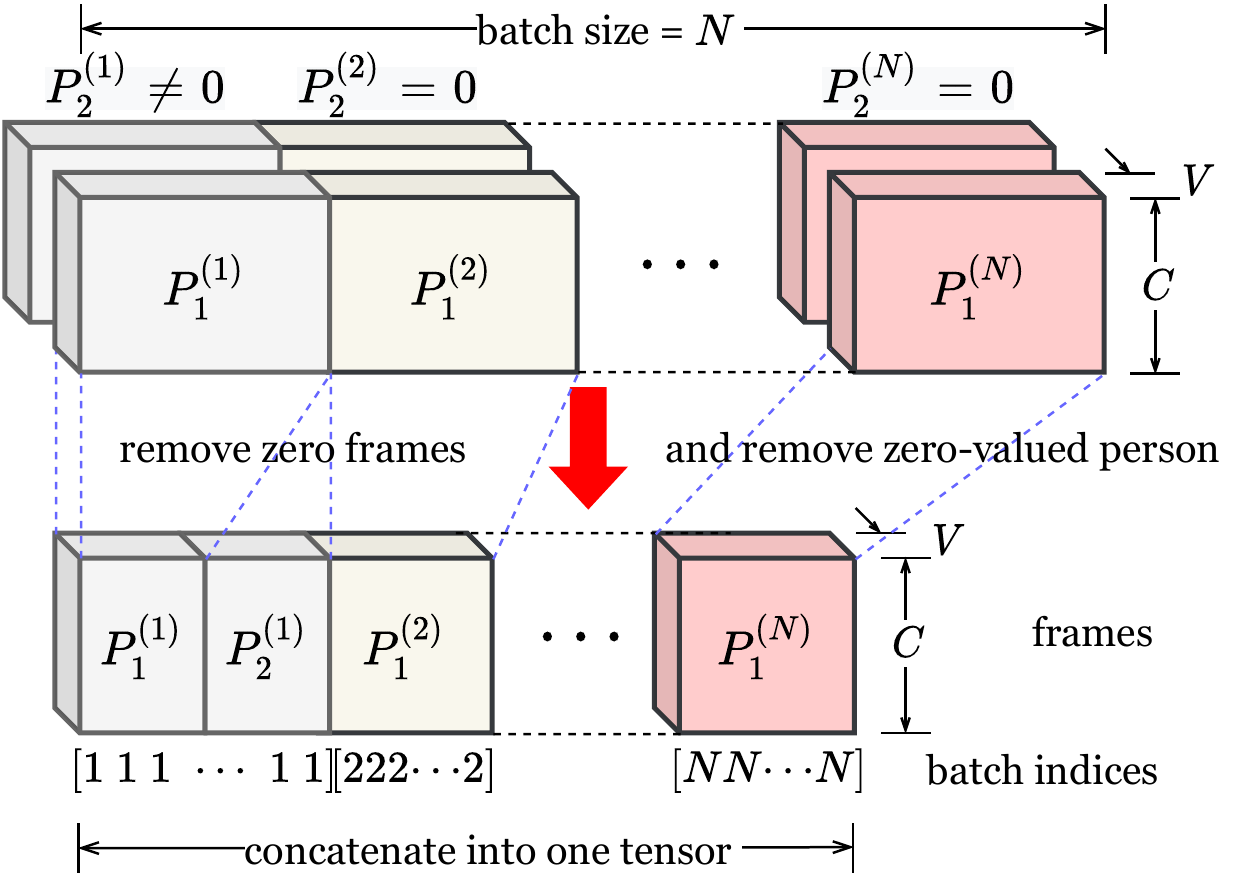}
    \caption{Illustration of our data format used in our framework: Previous works used the upper data format, which has fixed-sized time and person dimensions. Our work adopts new data format on the bottom, which has combined batch, person, and time dimensions into a single  variable length sequence.}
    \label{fig:data_format}
\end{figure}

The most apparent drawbacks in the previous approaches \cite{yan2018spatial, 2sagcn2019cvpr} are utilizing (1) the fixed number of frames for each video clip and (2) zero-filling for the non-existing second person. The first drawback constrains their scalability to process video clips longer than the predefined length and their flexibility on a shorter video clip. The second drawback due to the zero's participation in computation causes latency degradation. Moreover, a significant amount of memory space is allocated to those zero-valued data during the computation. So we propose a compact data format to bypass these drawbacks. Also, we propose Segmented Linear MHSA to process our compact data format.

\begin{figure*}[ht]
    \centering
    \includegraphics[width=0.8\textwidth]{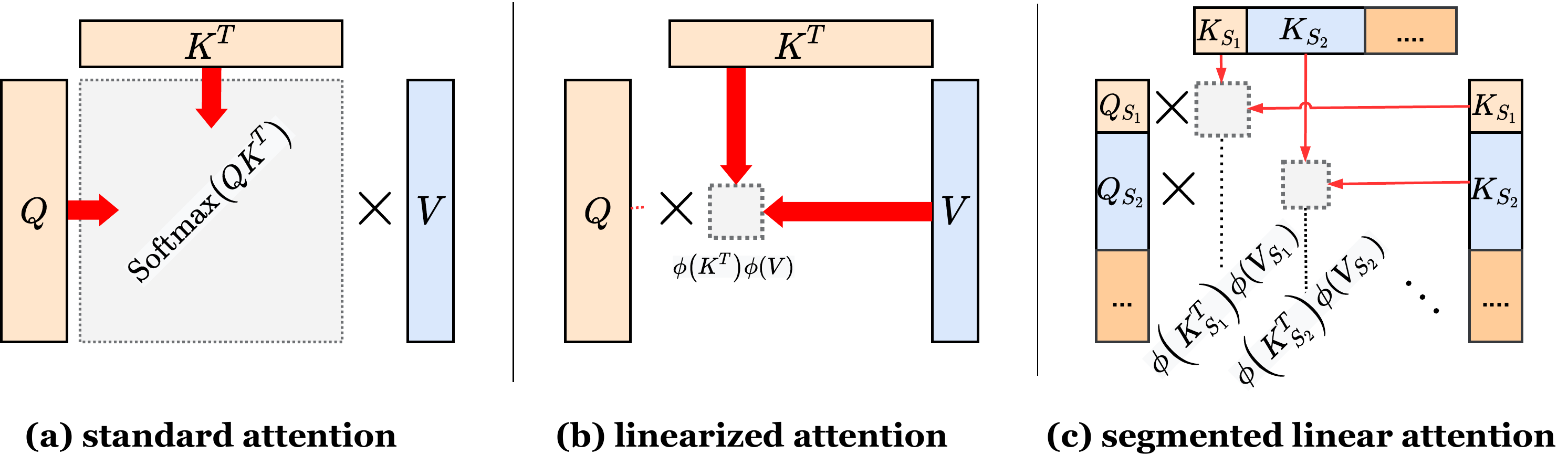}
    \caption{Illustration of Different Attention Operations: (a) Standard attention is obtained from $Softmax(QK^T)V$ with a complexity of $\mathcal{O}(n^2)$. (b) Linearized attention $\phi(Q) (\phi(K^T) V)$ with kernel function $\phi(\cdot)$ reduces the complexity to $\mathcal{O}(n)$, (c) we extend the linearized attention (b) to process segments of sequences.}
    \label{fig:attn_ops}
\end{figure*}

\subsubsection{Variable Frame Length Data Format} 
The Figure \ref{fig:data_format} shows the comparison between our data format and the format used by previous works. In the data format adopted by previous works, longer videos are cut off to the predefined length and shorter videos are padded with repeated frames. Furthermore, the frames with a single person are all zero-padded to match the fixed number of persons. The upper data format from Figure\ref{fig:data_format} illustrates the NTU RGB+D data format used by previous works. In each fixed-length video $V^{(i)}$, $P_1^{(i)}$ and $P_2^{(i)}$ represent two persons. In NTU RGB+D 120 dataset, only 26 out of 120 actions are mutual actions, which means that the second person's skeleton is just zeros ($P_2^{(i)} = 0$ in Figure \ref{fig:data_format}) in most data samples. In contrast to the previous data format, the proposed format maintains the original length of each video clip. Additionally, when a video clip contains two persons, we concatenate them along the frame dimension. Instead of keeping an individual dimension for a batch of video clips, we further concatenate the video clips in a batch along the frame dimension, and the auxiliary vector stores the batch indices to indicate to which video clip a frame belongs, as shown in the bottom data format of Figure \ref{fig:data_format}. Moreover, given the new dimensions ($N$, $V$, $C$) as shown in Figure \ref{fig:data_format}, where $N$ is the total number of frames after concatenating the video clips along the temporal dimension and $V$ is the number of skeleton's joints, we regard dimension $N$ as the logical batch size for spatial attention and dimension $V$ as the logical batch size for temporal attention.

\begin{figure}[ht]
    \centering
    \includegraphics[width=0.45\textwidth]{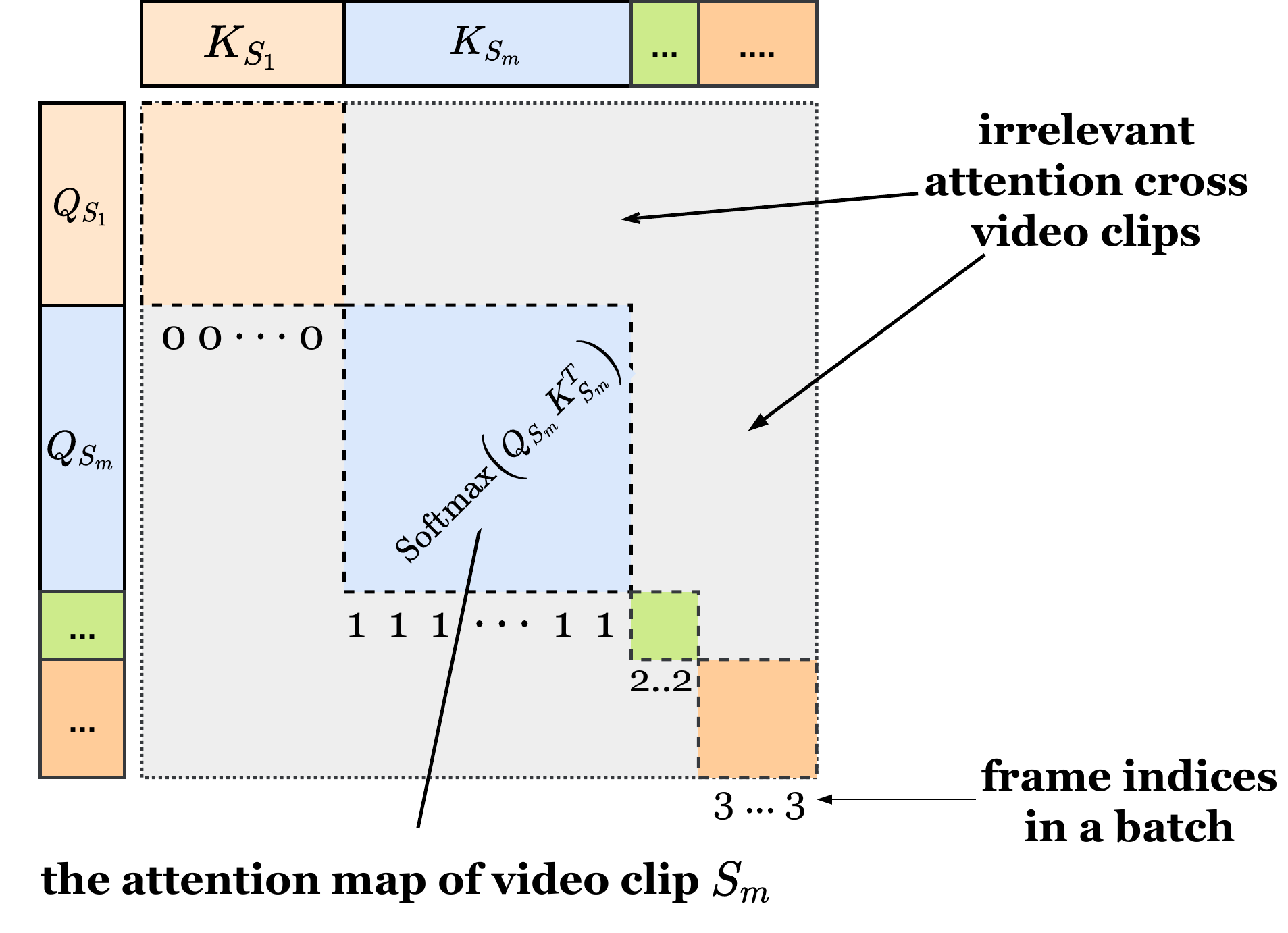}
    \caption{Segmented Attention: directly applying the linearized attention to our new data format will calculate unexpected attention between the two irrelevant video clips, which is error-prone. Therefore, we use segmented attention corresponding to each video sequence.}
    \label{fig:sgm_attn}
\end{figure}

\subsubsection{Segmented Linear Attention} \label{linear_attn}
With the new data format introduced in the previous section, we propose a novel linear multi-head attention tailored for this data format. We call it a Segmented Linear Attention. As stated in the previous sections, Transformers are originally designed for sequential data. In the human skeleton sequence, each joint's dynamic movement across the frames can be regarded as a time series. Therefore, the 3D coordinates, i.e., $(x, y, z)$, of every joint can be processed individually through the trajectory along the time dimension, and the application of attention extracts the interaction among time steps represented by frames. 

\textbf{Linear Attention}. Standard dot product attention mechanism \cite{attn2017all} (Equation \ref{eq:std_attn}) with the global receptive field of $N$ inputs are prohibitively slow due to the quadratic time and memory complexity $\mathcal{O}(N^2)$. The quadratic complexity also makes Transformers hard to train and limits the context. Recent research toward the linearized attention mechanism derives the approximation of the \textit{Softmax}-based attention. The most appealing ones are linear Transformers \cite{katharopoulos20a, choromanski2021rethinking, shen2021efficient} based on kernel functions approximating the \textit{Softmax}. The linearized Transformers can improve inference speeds up to three orders of magnitude without much loss in predictive performance \cite{tay2020efficient}.
Given the projected embeddings $Q$, $K$, and $V$ for input tensors of queries, keys, and values, respectively, according to the observation from the accumulated value $V_i' \in \mathbf{R}^{d}$ for the query $Q_i \in \mathbf{R}^{d}$ in position $i$, $d$ is the channel dimension, the linearized attention can be transformed from Equation \ref{eq:std_attn} to Equation \ref{eq:lnr_attn}, the computational complexity is reduced to $\mathcal{O}(Nd)$, when $d$ is much smaller than $N$, the computational complexity is approaching linear $\mathcal{O}(N)$:
\begin{equation}
    V_i' = \frac{\sum_{j=1}^N \left< Q_i, K_j \right> V_j}{\sum_{j=1}^N \left< Q_i, K_i \right>}
    \label{eq:std_attn}
\end{equation}

\begin{equation}
    \begin{split}
        V_i' &= \frac{\phi(Q_i)^T \sum_{j=1}^N \phi(K_j) V_j^T}{\phi(Q_i)^T \sum_{j=1}^N \phi(K_j)} = \frac{\phi(Q_i)^T U}{\phi(Q_i)^T Z} \\ 
        U &= \sum_{j=1}^N \phi(K_j) V_j^T, \quad Z = \sum_{j=1}^N \phi(K_j)
    \end{split}
    \label{eq:lnr_attn}
\end{equation}

where $\phi(\cdot)$ is the kernel function. In work of \cite{katharopoulos20a}, kernel function is simply simulated with ELU, $\phi(x) = elu(x) + 1$; while \cite{choromanski2021rethinking} introduces the \textit{Fast Attention via Orthogonal Random Feature} (FAVOR) maps as the kernel function, $\phi(x) = \frac{c}{\sqrt{M}} f(W x + b)^T$, where $c > 0$ is a constant, and $W \in \mathbf{R}^{M \times d}$ is a Gaussian random feature matrix, and $M$ is the dimensionality of this matrix that controls the number of random features.

\begin{figure*}[h!] 
    \centering
    \includegraphics[width=0.95\textwidth]{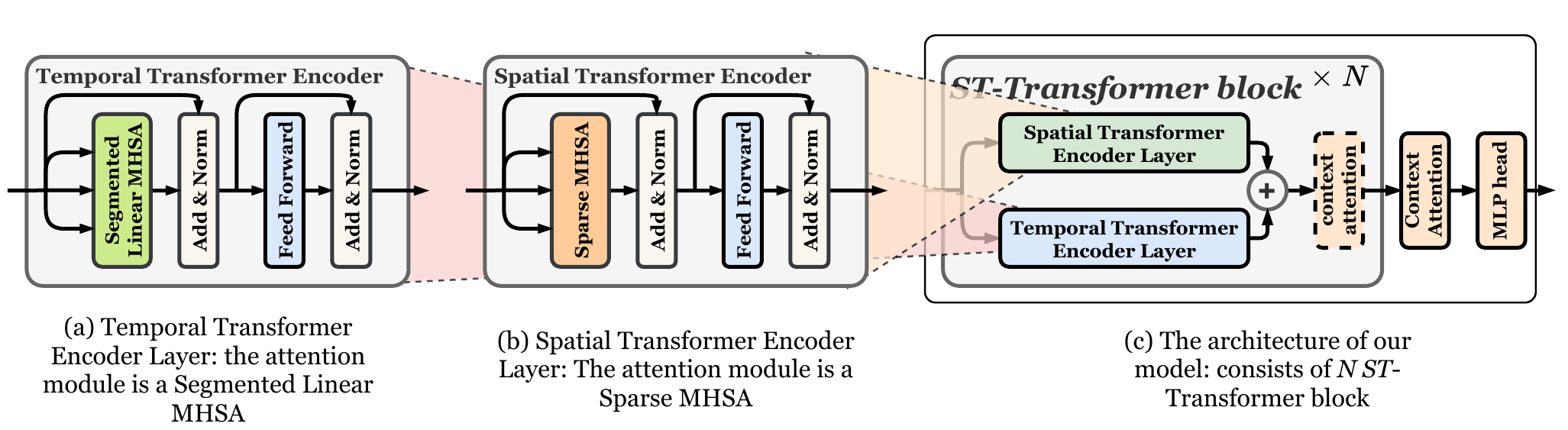}
    \caption{Illustration of the overall pipeline of our approach (STAR)}
    \label{fig:arch}
\end{figure*}

\textbf{Segmented Linear Attention}. Since we concatenate the various length of video clips within a single batch along the time dimension, directly applying linear attention will cause the cross clip attention, leading to irrelevant information taken into account from one video clip to another, as shown in Figure \ref{fig:sgm_attn}. Therefore, we consider the frames of a video clip arranged as a segment, and then we design the segmented linear attention by reformulating Equation \ref{eq:lnr_attn} with segment index. Therefore, for each $V_i$ in segment $\mathcal{S}_m$, we summarize
\begin{equation}
    \begin{split}
        V_{i \in \mathcal{S}_m}' &= \frac{\phi(Q_{i \in \mathcal{S}_m})^T \sum_{j \in \mathcal{S}_m} \phi(K_{j}) V_j^T}{\phi(Q_{i \in \mathcal{S}_m})^T \sum_{j \in \mathcal{S}_m} \phi(K_j)} \\ 
        &= \frac{\phi(Q_{i \in \mathcal{S}_m})^T U_{S_m}}{\phi(Q_{i \in \mathcal{S}_m})^T Z_{S_m}} \\
        U_{S_m} &= \sum_{j \in \mathcal{S}_m} \phi(K_{j}) V_j^T, \quad Z_{S_m} = \sum_{j \in \mathcal{S}_m} \phi(K_j) 
    \end{split}
    \label{eq:sgm_lin_attn}
\end{equation}

where $\mathcal{S}_m$ is the $m$-th segment, and the reduction operation $\sum_{j \in \mathcal{S}_m} f(x)$ can be easily implemented through the indexation to segments; and with help of the gathering and scattering operations \cite{torch_scatter}, the segmented linear attention maintains the highly-paralleled computation. Figure \ref{fig:attn_ops} illustrates the comparison of different attention operations.

\subsection{STAR Framework} \label{sec:arch}

In this work, we propose the Sparse-Transformer Action Recognition (STAR) framework. Figure \ref{fig:arch} (c) shows the overview of our STAR framework. The STAR framework is built upon several Spatial-Temporal Transformer blocks (ST-block) followed by context-aware attention and MLP head for classification. Each ST-block comprises two pipelines: the spatial Transformer encoder and the temporal Transformer encoder. Each Transformer encoder consists of several key components, including the \textbf{\textit{multi-head self-attention}} (MHSA), \textbf{\textit{skip connection}} (AND \& Norm part in Figure \ref{fig:arch} (c)), and \textbf{\textit{feed-forward network}}. The spatial Transformer encoder utilizes sparse attention to capture the topological correlation of connected joints for each frame. The temporal Transformer encoder utilizes the segmented linear attention to capture the correlation of joints along the time dimension. The output sum from the two encoder layers is fed to the context-aware attention module to perform weighted summarization on the sequence of frames. Positional encoding is also utilized before ST-block to provide the context of ordering on the input sequence. Below is a brief introduction to each of them.

\subsubsection{Context-aware attention} \label{sec:context_attn}
\begin{figure}[ht]
    \centering
    \includegraphics[width=0.45\textwidth]{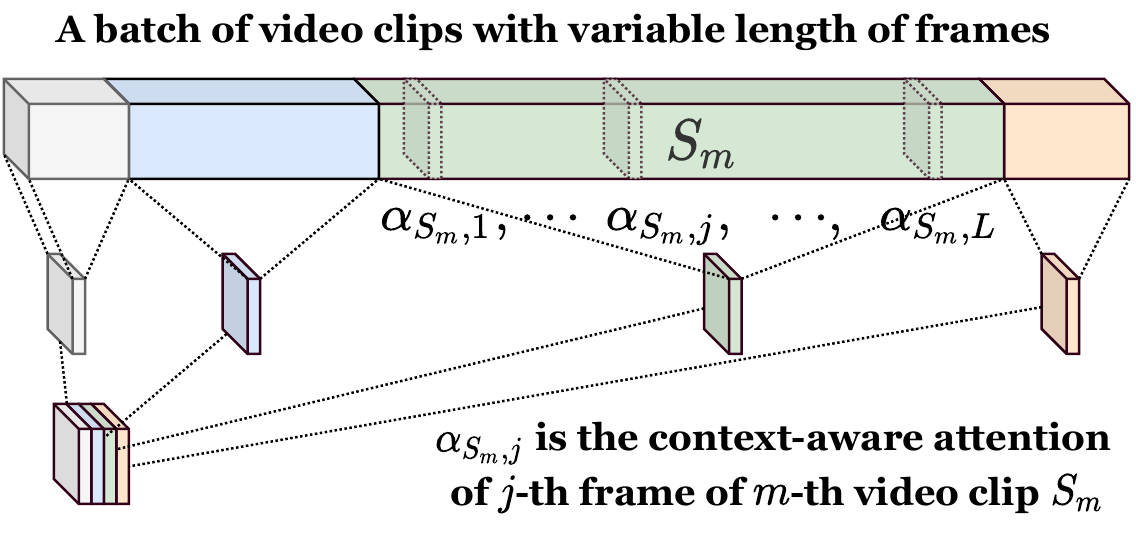}
    \caption{The context-aware attention is utilized to summarize each video clip.}
    \label{fig:context_attn}
\end{figure}

In previous works \cite{yan2018spatial, 2sagcn2019cvpr}, before connecting to the final fully-connected layer for classification, summarizing the video clip embedding along the temporal dimension is implemented by global average pooling. Alternatively, we utilize a probabilistic approach through \textit{context-aware attention}, which is extended from the work of \cite{simgnn2019}, to enhance this step's robustness, as demonstrated in Figure \ref{fig:context_attn}. Denote an input tensor embedding of video clip $S_m$ as $V \in R^{F \times N \times D}$, for $F$ is the number of frames in video clip $S_m$, $N$ is the number of joints of skeleton, and each joint possessing $D$ features, where $v_i \in R^{N \times D}$ is the embedding of frame $i$ of $V$. First, a \textit{global context} $c \in R^{N \times D}$ is computed, which is a simple average of embedding of frames followed by a nonlinear transformation: $c = tanh \left(\frac{1}{F} W \sum^{F}_{i \in S_m} v_i \right)$, where $W \in R^{D \times D}$ is a learnable weight matrix. The context $c$ provides the global structural and feature information of the video clip that is adaptive to the similarity between frames in video clip $S_m$, via learning the weight matrix. Based on $c$, we can compute one attention weight for each frame. For frame $i$, to make its attention an aware of the global context, we take the inner product between $c$ and its embedding. The intuition is that, frames similar to the global context should receive higher attention weights. A sigmoid function $\sigma(x) = \frac{1}{1 + exp(-x)}$ is applied to the result to ensure the attention weights is in the range $(0, 1)$. Finally, the video clip embedding $v' \in R^{N \times D}$ is the weighted sum of video clip embeddings, $v' = \sum^{F}_{i \in S_m} a_i v_i$. The following equations summarize the proposed context-aware attentive mechanism:
\begin{equation}
    \begin{split}
        c &= tanh \left( \frac{1}{F} W \sum^{F}_{j \in S_{m}} v_j \right) = tanh \left( \frac{1}{F} \left( V^T \cdot \mathbf{1} \right) W \right) \\
    v' &= \sum^{F}_{i \in S_m} \sigma \left( v_i^{T} \left[ tanh \left(
\frac{1}{F} W \sum^{F}_{j \in S_{m}} v_j \right) \right] \right) v_i \\
      &= \sum^{F}_{i \in S_{m}} \sigma \left( v_i^{T} c \right) v_i = \left[ \sigma (V c) \right]^T V    
    \end{split}
\end{equation}

\subsubsection{Positional Encoding} \label{sec:pos_enc}
As the attention mechanism is order-agnostic to the permutation in the input sequence \cite{attn2017all, tsai2019TransformerDissection} and treats the input as an unordered \textit{bag} of element. Therefore, an extra positional embedding is necessary to maintain the data order, i.e., time-series data are in the inherently sequential ordering. Then these positional embedding are participating the evaluation of the attention weight and value between token $i$ and $j$ in the input sequence.

\begin{figure}[ht]
    \centering
    \includegraphics[width=0.47\textwidth]{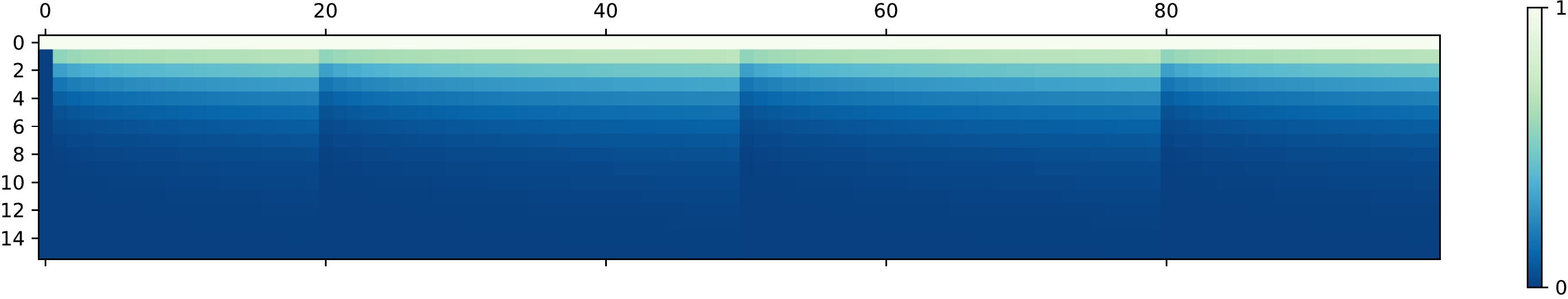}
    \caption{Illustration of Segmented Positional Encoding for a batch of 4 video clips. x-axis represents the number of frames and y-axis represents the feature dimension.}
    \label{fig:sgm_pos_enc}
\end{figure}

\textbf{Segmented Sequential Positional Encoding}
However, as we arrange the variable-length video clips into a batch along the temporal dimension, it is not feasible to directly apply positional encoding to the whole batch. Therefore, we introduce the \textit{segmented positional encoding} where each video clip gets its positional encoding according to batch indices. An example of such encoding is shown in Figure \ref{fig:sgm_pos_enc}.

\textbf{Structural Positional Encoding}. we also attempt to apply the structural positional encoding, e.g., tree-based positional encoding \cite{neurips2019treepositional, omote-etal-2019-dependency}, to the spatial dimension, i.e., the tree topology of skeleton. Experiments show that the current approach which we used cannot improve our model's performance significantly. Hence, to reduce our model's complexity, we decide not to apply the structural positional encoding for this work and leave it for future research.

\begin{table*}[h!]
\centering
\begin{tabular}{ccccc}
       & \multicolumn{2}{c}{NTU-60}  & \multicolumn{2}{c}{NTU-120} \\ 
\toprule \textbf{Method} & \textbf{X-subject} & \textbf{X-view} & \textbf{X-subject} & \textbf{X-setup} \\ 
\hline
ST-GCN & 81.5 & 88.3 & 72.4& 71.3\\
ST-TR & 88.7 & 95.6 & 81.9.& 84.1\\
STAR-64 (ours) & 81.9 & 88.9 & 75.4& 78.1\\ 
STAR-128 (ours) & 83.4 & 89.0 & 78.3& 80.2\\
\hline
\end{tabular}
\caption{Comparison of models' accuracy on NTU RGB+D 60 and 120 datasets}
\label{tab:accuracy}
\end{table*}

\begin{table*}[!h]
\centering
\begin{tabular}{cccc}
\hline
\textbf{Model}    & \textbf{CUDA time (ms)} & \textbf{num. of parameters} & \textbf{GMACs}\\ \hline
ST-GCN   & 333.89  & 3.1M               & 261.49\\
ST-TR    & 1593.05  & 6.73M              & 197.55\\
STAR-64 (ours)  & 86.54   & 0.42M              & 15.58\\
STAR-128 (ours) & 191.23    & 1.26M             & 73.33\\ \hline
\end{tabular}
\caption{Comparison of models' efficiency}
\label{tab:efficiency}
\end{table*}

\section{Experiments}
In this section, we conduct experiments and ablation studies to verify the effectiveness and efficiency of our proposed sparse spatial and segmented linear temporal self-attention operations. The comparison has been made with \textbf{ST-GCN}, the baseline GCN model, and \textbf{ST-TR}, one of the state-of-the-art hybrid model, which have utilized full attention operation coupled with graph convolutions. The corresponding analysis demonstrates the potential of our model and the possible room for improvements.

\subsection{Datasets}
In the experiments, we evaluate our model on two largest scale 3D skeleton-based action recognition datasets, NTU-RGB+D 60 and 120. 
\subsubsection{NTU RGB+D 60}
This dataset contains 56,880 video clips involving 60 human action classes. The samples are performed by 40 volunteers and captured by three Microsoft Kinect v2 cameras \cite{shahroudy2016ntu}. It contains four modalities, including RGB videos, depth sequences, infrared frames, and 3D skeleton data. Our experiments are only conducted with the 3D skeleton data. The length of the action samples vary from 32 frames to 300 frames. In each frame, there are at most 2 subjects and each subject contains 25 3D joint coordinates. The dataset follows two evaluation criteria, which are Cross-Subject and Cross-View. In the Cross-View evaluation (X-View), there are 37,920 training samples captured from camera 2 and 3 and 18,960 test samples captured from camera 1. In the Cross-Subject evaluation (X-Sub), there are 40,320 training samples from 20 subjects and 26,560 test samples from the rest. We follow the original two benchmarks and report the Top-1 accuracy as well as the profiling metrics.
\subsubsection{NTU RGB+D 120}
The dataset \cite{liu2019ntu} extends from NTU RGB+D 60 and is currently the largest dataset with 3D joint annotations. It contains 57,600 new skeleton sequences representing 60 new actions, a total of 114,480 videos involving 120 classes of 106 subjects captured from 32 different camera setups. The dataset follows two criteria, which are Cross-Subject and Cross-Setup. In the Cross-Subject evaluation, similar to the previous dataset, splits subjects in half to training and testing dataset. In the Cross-Setup evaluation, the samples are divided by the 32 camera setup IDs, where the even setup IDs are for training and the odd setup IDs for testing. Similar to the previous dataset, there is no preprocessing to set the uniform video length for all the samples. We follow the two criteria and report the Top-1 accuracy and the profiling metrics.

Unlike GCN-based models, where the length of all the samples and the number of subjects need to be fixed (e.g. 300 frames and 2 subjects), our model can process varying length of input samples and of the number of subjects. So no further preprocessing with padding is done on the samples.

\subsection{Configuration of experiments}
\textbf{Implementation details}. As the original Transformer framework \cite{attn2017all} employs the unified model size $d$ for every layer, we follow the same notion and keep the hidden channel size uniform across the attention heads and the feedforward networks. We run the experiments with two different hidden channel sizes, 64 and 128 for our Transformer encoders (STAR-64 and STAR-128), respectively. The hidden channel size of the MLP head is also proportional to that of the attention heads. Our model consists of 5 layers, each layer comprises one spatial encoder and one temporal encoder in parallel and the  output sum from the two encoders is fed to the next layer. Drop rates are set to 0.5 for every module. We also replace the ReLU non-linear activation funciton with SiLU (or Swish) \cite{elfwing2018sigmoid, swish2017} to increase the stability of gradients in back-propagation phase (GELU or SELU also bring similar effect). Our model is implemented with the deep learning framework PyTorch \cite{pytorch} and its extension PyTorch Geometric \cite{pytorch_geometric_Fey2019}. The scattering/gathering operations and sparse matrix multiplications are based on PyTorch Scatter \cite{torch_scatter} and PyTorch Sparse \cite{torch_sparse}, respectively.

\begin{table*}[!h]
\centering
\begin{tabular}{cccc}
    & \textbf{MACs} & \textbf{Parameters} & \textbf{Latency} \\ \toprule 
    ST-GCN & 
    \begin{tabular}[l]{@{}l@{}}
       \textit{Conv2d}: 260.4 GMACs \\ 
       \textit{BatchNorm2d}: 737.3 MMACs \\ 
       \textit{ReLU}: 184.3 MMACs
    \end{tabular} & 
    \begin{tabular}[l]{@{}l@{}}
       \textit{Conv2d}: 3.06M \\ 
       \textit{BatchNorm2d}: 6.4K \\ 
       \textit{Linear}: 15.4K 
    \end{tabular} & 
    \begin{tabular}[l]{@{}l@{}}
       \textit{Conv2d}: 149.92ms \\ 
       \textit{BatchNorm2d}: 19.92ms \\ 
       \textit{ReLU}: 4.49ms
    \end{tabular} \\ \hline

    ST-TR  & 
    \begin{tabular}[l]{@{}l@{}}
       \textit{Conv2d}: 810.57 GMACs \\ 
       \textit{MatMul}: 161.1 GMACs \\ 
       \textit{BatchNorm2d}: 138.4 MMACs
    \end{tabular}  & 
    \begin{tabular}[l]{@{}l@{}}
       \textit{Conv2d}: 2.7M \\ 
       \textit{BatchNorm2d}: 10.5K \\ 
       \textit{Linear}: 30.8K
    \end{tabular} & 
    \begin{tabular}[l]{@{}l@{}}
       \textit{Conv2d}: 692.39ms \\ 
       \textit{MatMul}: 161.38ms \\ 
       \textit{BatchNorm2d}: 38.97ms
    \end{tabular} \\ \hline

    STAR-64   & 
    \begin{tabular}[l]{@{}l@{}}
       MatMul(attention): 24.4 GMACs\\
       Mul(sparse): 12.3 GMACs \\
       Linear: 6.2 GMACs
    \end{tabular} & 
    \begin{tabular}[l]{@{}l@{}}
    Linear:83.2K\\
    LayerNorm: 1.3K
    \end{tabular} & 
    \begin{tabular}[l]{@{}l@{}}
       MatMul: 25.27ms \\
       Mul: 12.81ms\\
       Linear:6.53ms
    \end{tabular}   \\ \hline

\end{tabular}
\caption{The breakdown analysis and top-3 components in each metrics}
\label{tab:3_metrics}
\end{table*}

\textbf{Training setting}. The maximum number of training epochs is set to 100. We used the Adam optimizer \cite{kingma2014adam} with $\beta_1 = 0.9$, $\beta_2 = 0.98$ and $\epsilon = 10^{-9}$. Following the setting of the original Transformer paper \cite{attn2017all}, the learning rate is adjusted throughout the training:
\begin{equation}
    lrate = d^{-0.5} \cdot min(t^{-0.5}, t \cdot w^{-1.5})
    \label{eq:warmup}
\end{equation}
where $d$ is the model dimension, $s$ is \textit{step number}, and $w$ is the \textit{warmup steps}. According to the equation \ref{eq:warmup}, the learning rate linearly increases for the first $w$ training steps and then decreases proportionally to the inverse square root of the step number. We keep the original settings for the baseline models in their papers \cite{yan2018spatial, plizzari2020spatial}, and use their codes provided online. 
All our training experiments are performed on a system with two GTX TITAN X GPUs and a system with one TITAN RTX GPU, while the inferences are executed on a single GPU. 

\subsection{Results and Analysis}
We evaluate the accuracy and the efficiency of the baseline GCN model (ST-GCN), our model (STAR) and the hybrid model (ST-TR), which utilize both transformer and GCN frameworks.

\subsubsection{Accuracy}
\begin{figure*}[h!]
    \centering
    \includegraphics[width=.76\textwidth]{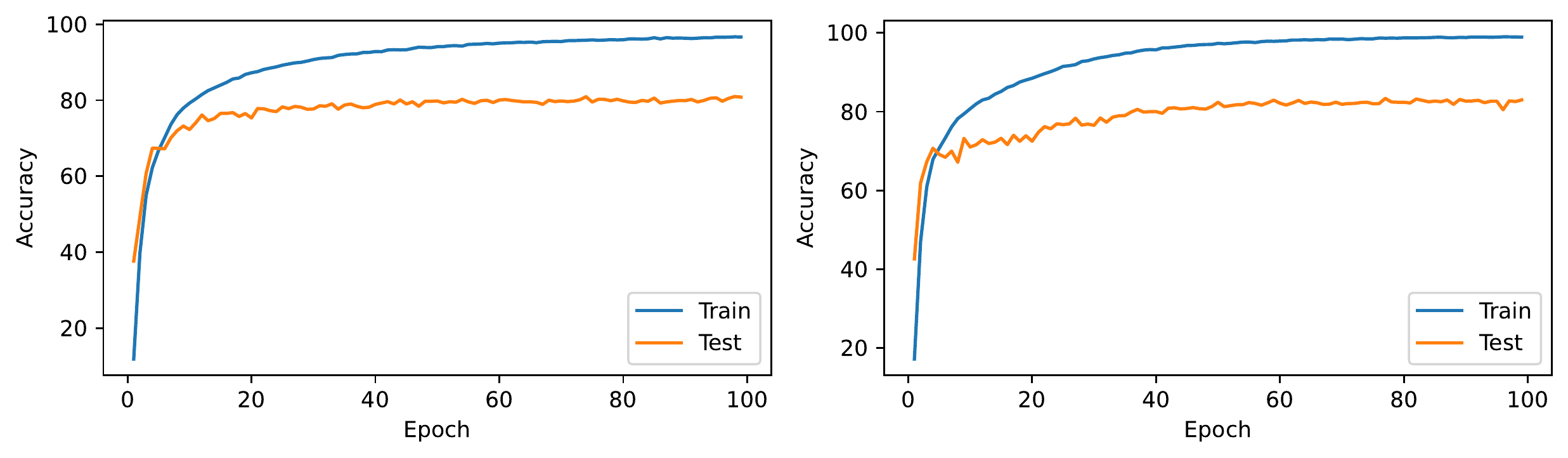}
    \caption{The training and testing curve for STAR-64(left) and STAR-128(right) on NTU-RGB+D 60 X-subject benchmark.}
    \label{fig:general}
\end{figure*}

We first evaluate the effectiveness of our Transformer encoder based model compared to ST-TR and ST-GCN models. Each model's accuracy is evaluated with the NTU RGB+D 60 and 120 testing datasets. As shown in the Table \ref{tab:accuracy}, our model outperforms ST-GCN in both cross-view (cross-set) and cross-subject benchmarks of the two dataset. Our model achieves 3.6 $\sim$ 7.7 percent lower accuracy compared to ST-TR, which heavily relies on convolution-based key components inherited from ST-GCN and utilizes them in both spatial and temporal pipelines. Our model yields modest performance compared to the state-of-the-art models in NTU RGB+D 60 and 120 when trained from scratch. The Figure \ref{fig:general} shows that there exists a performance gap between the training and testing. Transformer architectures' lack of inductive biases, especially translation equivariance and locality that are essential to convolution networks, could result in weak generalization. In NLP, Transformers are usually pretrained on a large corpus of text and fine-tuned on a smaller task-specific dataset to boost the performance. We would like to conduct extensive experiments on pre-training and fine-tuning our model on a larger dataset in the future to improve the accuracy comparable to those of the state-of-the-art models. For our future study, we want to address effective generalization methods for Transformer models, which resolves overfitting issues and improve the overall performance.

\subsubsection{Efficiency} 
In this section, we evaluate the efficiency of the different models. As shown in Table \ref{tab:efficiency}, our model (STAR) is significantly efficient in terms of model size, the number of multiply-accumulate operations (GMACs) and the latency. Each metric for the different models is evaluated by running inference with sample dataset. Our model is fed with the original skeleton sequence of varying length. The other two models are fed with fix-sized skeleton sequence padded to 300 frames and 2 persons. We use the official profiler of PyTorch (v1.8.1) \cite{pytorch}, and Flops-Profiler of DeepSpeed \cite{deepspeed2020kdd} to measure the benchmarks. The results are summarized with the following metrics:
\begin{figure*}[h!]
    \centering
    \includegraphics[width=0.84\textwidth]{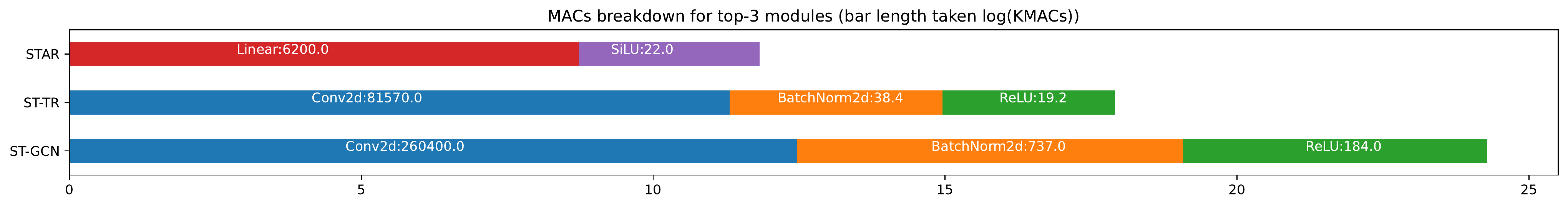}
    \caption{The breakdown of MACs for top-3 modules}
    \label{fig:mac_breakdown}
\end{figure*}

\begin{figure*}[h!]
    \centering
    \includegraphics[width=0.84\textwidth]{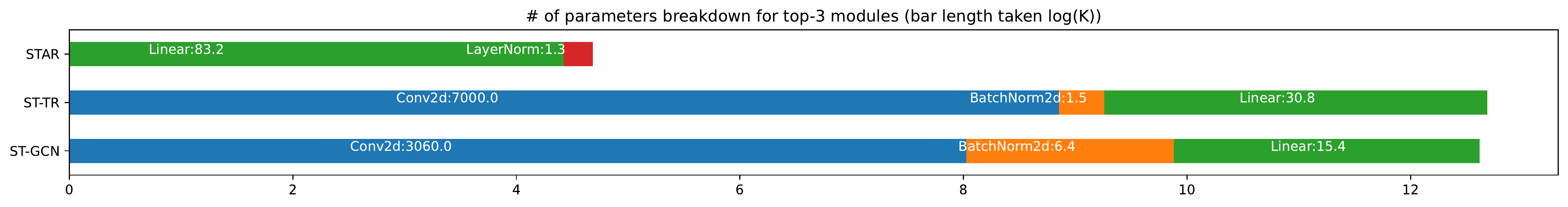}
    \caption{The breakdown of \# parameters for top-3 modules} \label{fig:param_breakdown}
\end{figure*}

\begin{figure*}[h!]
    \centering
    \includegraphics[width=0.84\textwidth]{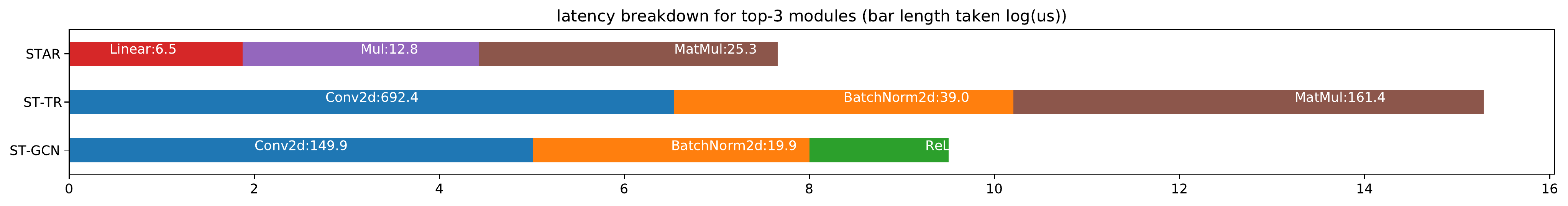}
    \caption{The breakdown of latency for top-3 modules}
    \label{fig:latency_breakdown}
\end{figure*}

\textbf{MACs}. The number of Multiply-Accumulate (MAC) operations is used to determine the efficiency of deep learning models. Each MAC operation is counted as two floating point operations. With the same hardware configuration, more efficient models require fewer MACs than other models to fulfill the same task. As shown in Table \ref{tab:efficiency}, both of our model with different channel sizes execute only $\frac{1}{3} \sim \frac{1}{17}$ amount of GMACs (i.e., Giga MACs) compared to ST-GCN and ST-TR models, respectively.

\textbf{Model size}. Model size is another metric to measure the efficiency of a machine learning model. Given the same task, smaller model delivering the same or very close performance is preferable. Smaller model is not only beneficial for the higher speedup and less memory accesses but also gives better energy consumption, especially for embedded systems and edge devices with scarce computational resources and small storage volume. The column of the number of parameters in Table \ref{tab:efficiency} depicts the size of the models, these parameters are trainable weights in the model. Among all the model, STAR possesses the smallest model size,  0.42M and 1.26M for STAT-64 and STAR-128, respectively.

\textbf{Breakdown analysis}.
The breakdown analysis is used to identify potential bottlenecks within different models (STAR-64, ST-TR, and ST-GCN). Table \ref{tab:3_metrics} provides the detailed profiling results for the top-3 computation modules that are dominant in each models. According to Figure \ref{fig:mac_breakdown}, \ref{fig:param_breakdown} and \ref{fig:latency_breakdown}, the convolution operations cost significant number of MAC operations and lead to computation bound. ST-GCN and ST-TR mainly consist of the convolution operations followed by batch normalization, which requires relatively large computational resources. Our Transformer model is based on sparse and linear attention mechanisms. It only produces relatively small attention weights from sparse attention; and performs low-rank matrix multiplication for linear attention ($\mathcal{O}(n)$). This replaces huge dynamic weights of attention coefficients from the standard attention mechanism, which has a quadratic time and space complexity ($\mathcal{O}(n^2)$). 

\subsection{Ablation Study}.
In this section, we evaluate the effectiveness and efficiency of our sparse self-attention operation in spatial encoder compared to the standard transformer encoder with full-attention operation. Table \ref{tab:acc_ablation} and Table \ref{tab:eff_ablation} show that our model with sparse self-attention operation achieves higher accuracy on both X-subject and X-view benchmarks and use significantly less number of GMACs and runtime. This shows that additional correlations of distant joints calculated by full attention do not improve the performance but rather contribute noise to the prediction. To handle such issue, learnable masks, consistent with adjacency matrix of skeleton, can be integrated to the full attention calculation to avoid accuracy degradation. But it requires extra computation involving learnable masks. 

\begin{table}[h!]
\centering
\begin{tabular}{ccccc}
       & \multicolumn{2}{c}{NTU-60}  & \multicolumn{2}{c}{NTU-120} \\ 
\toprule Method & X-subject & X-view & X-subject & X-setup \\ 
\hline
    STAR (sparse) & 83.4&84.2 & 78.3& 78.5 \\
    STAR (full) & 80.7&81.9 & 77.4& 77.7 \\
\hline
\end{tabular}
    \caption{Classification accuracy comparison between Sparse attention and Full attention on the NTU RGB+D 60 Skeleton dataset.}
\label{tab:acc_ablation}
\end{table}

\begin{table}[!h]
\centering
\begin{tabular}{ccccc}
\hline
Model    & CUDA time (ms) & GMACs  \\ \hline
STAR-sparse  & 105.7   & 15.58 \\
STAR-full & 254.7      & 73.33    \\ \hline
\end{tabular}
\caption{Efficiency comparison between Sparse attention and Full attention on the NTU RGB+D 60 Skeleton dataset.}
\label{tab:eff_ablation}
\end{table}


\section{Conclusion}
In this paper, we propose an efficient Transformer-based model with sparse attention and segmented linear attention mechanisms applied on spatial and temporal dimensions of action skeleton sequence. We demonstrate that our model can replace graph convolution operations with the self-attention operations and yield the modest performance, while requiring significantly less computational and memory resources. We also designed compact data representation which is much smaller than fixed-size and zero padded data representation utilized by previous models. This work was supported in part by Semiconductor Research Corporation (SRC).
 

\bibliography{references.bib}

\end{document}